\documentclass{new_tlp}
\usepackage{mathptmx}
\usepackage{hyperref}
\usepackage{graphicx}
\usepackage{picinpar}
\usepackage{xcolor}
\usepackage{amsmath}

\newcommand\bcmdtab{\noindent\bgroup\tabcolsep=0pt%
  \begin{tabular}{@{}p{10pc}@{}p{20pc}@{}}}
\newcommand\ecmdtab{\end{tabular}\egroup}

\usepackage{picinpar}
\usepackage{times}
\usepackage{helvet}
\usepackage{courier}

 \usepackage{todonotes}
 
\usepackage{graphicx} 
\usepackage{algorithm}
\usepackage{algorithmic}
\usepackage{epstopdf}
\usepackage{url}

\usepackage{amsfonts, amsmath, amssymb, mathrsfs}
\usepackage{times, helvet, courier}
\usepackage{graphicx, epsfig, epstopdf}
\usepackage{wrapfig, caption, multirow, url}

\usepackage{algorithm}
\usepackage{algorithmic}
\usepackage[font=small,format=hang,parskip=1pt]{caption}

\usepackage{stmaryrd}
\usepackage{url}
\usepackage{csquotes}
\usepackage{listings}
\usepackage{algorithm}
\usepackage{algorithmic}
\usepackage{todonotes}

  \title[Creating Phylotastic]
        {Phylotastic: An Experiment in Creating, Manipulating, and Evolving
        	 Phylogenetic Biology Workflows Using Logic Programming}

  \author[T. Nguyen, E. Pontelli, T. C. Son]
         {THANH HAI NGUYEN, ENRICO PONTELLI, TRAN CAO SON\\
         	Department of Computer Science,
         	New Mexico State University\\
         \email{thanhnh | epontell | @nmsu.edu, tson@cs.nmsu.edu}}

\jdate{February 2018}
\pubyear{2018}
\pagerange{\pageref{firstpage}--\pageref{lastpage}}
\doi{S1471068401001193}

\allowdisplaybreaks

\begin{document}

\label{firstpage}

\maketitle

  \begin{abstract}
    Evolutionary Biologists have long struggled with the challenge of developing 
    analysis workflows in a flexible manner, thus facilitating the reuse of phylogenetic
    knowledge. An evolutionary biology workflow can be viewed as a plan which composes web
    services that can retrieve, manipulate, and produce phylogenetic trees.
    The Phylotastic project was launched two years ago as a collaboration between evolutionary
    biologists and computer scientists, with the goal of developing an open architecture to facilitate
    the creation of such analysis workflows. While composition of web services is a problem that
    has been extensively explored in the literature, including within the logic programming domain,
    the incarnation of the problem in Phylotastic provides a number of additional challenges. 
    Along with the need to integrate preferences and formal ontologies in the description of the
    desired workflow, evolutionary biologists tend to construct workflows in an incremental manner,
    by successively refining the workflow, by indicating desired changes (e.g., exclusion of
    certain services, modifications of the desired output). This leads to the need of 
    successive iterations of incremental replanning, to develop a new workflow that integrates
    the requested changes while minimizing the changes to the original workflow. 
  This paper illustrates how Phylotastic has addressed the challenges of creating and refining
  phylogenetic analysis workflows using logic programming technology and how such solutions have
  been used within the general framework of the Phylotastic project. 
  \end{abstract}

  \begin{keywords}
    Bioinformatics, workflows, web services, planning, composition, re-composition, similarity, quality of 
    service
  \end{keywords}

\section{Introduction}

A \emph{phylogeny} (phylogenetic tree) is a representation of the evolutionary history of a set of entities---in the context of
this work, we focus on phylogenies describing biological entities (e.g., organisms). Typically, 
a phylogeny is a branching diagram showing relationships between species, but phylogenies can be drawn for individual 
genes, for populations, or for other entities (e.g., non-biological applications of phylogenies include the study
of evolution of languages). Phylogenies are built by analyzing specific properties of the species
(i.e., \emph{characters}), such as morphological traits (e.g., body shape, placement of bristles or shapes of cell
walls), biochemical, behavioral or molecular features of species or other groups. In building a tree, species are organized into nested groups based on
shared derived traits (traits different from those of the group's ancestor). Closely related species typically have fewer 
differences among the value of their characters,  while less related 
species tend to have more.
Currently, phylogenetic trees can be either explicitly constructed (e.g., from a collection of descriptions of species), or extracted from repositories phylogenies, such as {\small \tt OpenTree} \footnote{{\small \url{http://tree.opentreeoflife.org}}} and {\small \tt TreeBASE} \footnote{{\small \url{https://treebase.org}}}. In a phylogeny, the topology is the branching structure of the tree. It is of  biological significance, because it indicates patterns of relatedness among {\small\tt taxa}, meaning that trees with the same topology provide the same biological interpretation. Branches show the path of transmission from one generation to the next. Branch lengths indicate genetic change, i.e., the longer the branch, the more genetic change (or divergence) has occurred. A variety of methods have been devised to estimate a phylogeny from the traits of the taxa (e.g., \cite{felse}).

 Phylogenies are
useful in all areas of biology,  to provide a hierarchical framework for classification and for
process-based models that allow scientists to make robust inferences from comparisons of evolved
things. 
A standing dream in the field of evolutionary biology is the assembling a \emph{Tree of Life (ToL),} 
a phylogeny broadly covering some
$10^7$ species \cite{ph1,ph2}. The first draft of a grand phylogenetic synthesis---a single tree with perhaps $2.5 \times 10^6$ 
species---is emerging  from the Open Tree of Life (OpenTree) project. Yet, the current state of the ToL is a
collection of trees, and there are various reasons to expect that this situation will persist. When we
refer to ``the ToL'' here, we mean the dispersed set of available species trees, with a strong focus on
larger trees from reputable published sources (e.g., \cite{ph4,birds,ph5}).

While experts continue expanding, consolidating, and improving the ToL, our motivation is to put this
expert knowledge into the hands of everyone: ordinary researchers, educators, students, and the public.
To achieve this, we launched the \emph{Phylotastic} project, aimed at building a community-sustainable architecture to support
flexible on-the-fly delivery of expert phylogenetic knowledge.
The premise of disseminating knowledge is that it will be re-used. How do trees get re-used? On a
per-tree basis, re-use is rare---most trees are inferred \emph{de novo} for a specific study, rendered as
images in a published report, stored on someone's hard drive, and (apparently) not used again \cite{ph3}.
Yet, large species trees are re-used in ways that small species trees (and sequence-family trees)
are not. In a sample of 40 phylogeny articles, \citeN{ph3} found 6 cases in which scientists
obtained a custom tree by extraction from a larger species tree. These studies implicate diverse
uses: functional analyses of leaf traits or lactation traits; phylogenetic
diversity of forest patches; analyzing niche-diversity correlations, spatial distribution of wood
traits, and spatial patterns of diversity. The implicated trees include those covering $4,510$
extant mammals, $55,473$ angiosperm species, and $1,566$ angiosperm taxa.

The Phylotastic project offers a solution to  the reuse of phylogenetic knowledge problem, by adopting a web services
composition approach. The \emph{phyloinformatic} community has been very active in the development of a diversity
of data repositories and software tools, to collect and analyze artifacts relevant to evolutionary analysis. These
tools are sufficient to realize all of the studies in the previously
mentioned papers, when properly integrated in a coherent workflow. 
Furthermore, the analysis protocols adopted in such studies are often
the result of an iterative process, where the protocol is successively refined to better suit the available
data and produce results of adequate quality \cite{ph3}.

In this paper, we describe the infrastructure used by Phylotastic to achieve the following goals:
{\bf (1)} Allow a user to provide the available knowledge about the desired protocol (e.g., input, type of
output, selected classes of operations that should be used); 
{\bf (2)} Derive a collection of workflows
that satisfy the desired conditions, through a web service discovery and composition process; 
{\bf (3)} Allow the user to suggest manipulations of a chosen workflow (e.g., exclusion of a service, addition
of another output); 
{\bf (4)} Determine new workflows that satisfy the requested manipulations while 
maintaining maximum similarity with the chosen workflow. All the manipulations of the workflows---i.e., 
composition of web services, modification of workflows, computation of similarity between workflows---are
realized in Answer Set Programming (ASP). ASP provides a clear advantage, allowing the simple integration
of different forms of knowledge (e.g., ontologies describing services and artifacts, user preferences) and
facilitating the encoding of the composition and re-composition problems as ASP planning problems. We present
the overall structure of the Phylotastic architecture, describe how web service composition is encoded in ASP,
and analyze how workflow refinements are achieved. We also demonstrate various aforementioned features 
through a use-case. 
 
\section{Background: The Phylotastic Project and Its Implementation}
 
\subsection{Architecture}

\begin{figure}[htbp]
  \centering \includegraphics[width=0.7\textwidth]{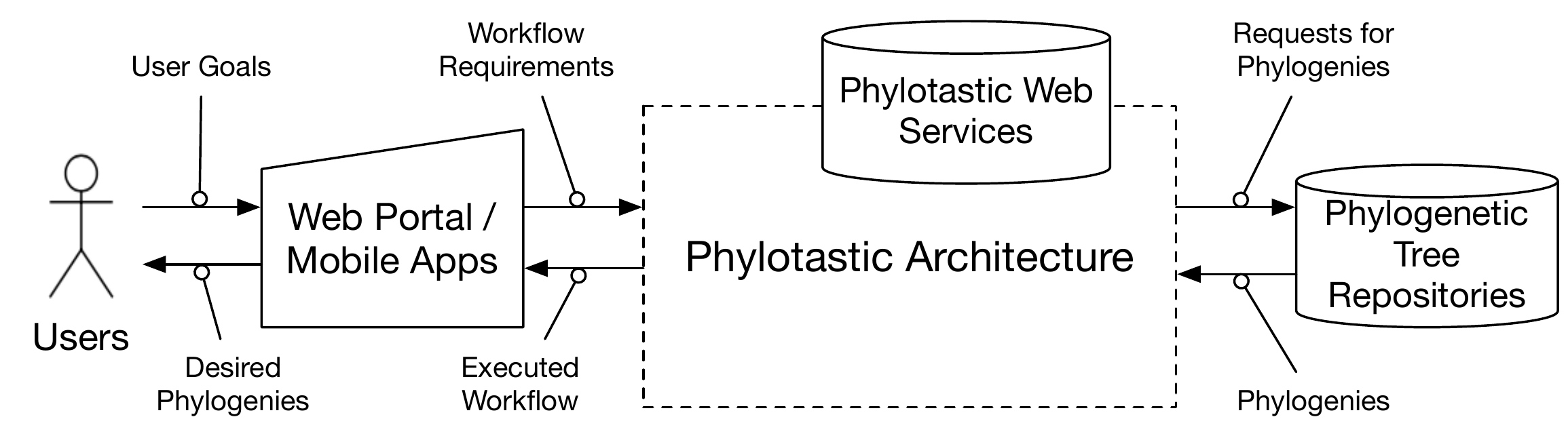}
  \caption{Overall Phylotastic Structure}
  \label{overall}
\end{figure}

The Phylotastic project proposes  a flexible system for on-the-fly delivery of custom 
phylogenetic trees that would support many kinds of tree re-use, and be open 
for both users and data providers. Phylotastic proposes an open architecture, composed of a collection of \emph{web services} relevant 
to creation, storage, and reuse of phylogenetic knowledge, that can be assembled in user-defined workflows through a Web portal and mobile applications (Fig. \ref{overall}).

\begin{figure}[htbp]
  \centering \includegraphics[width=0.9\textwidth]{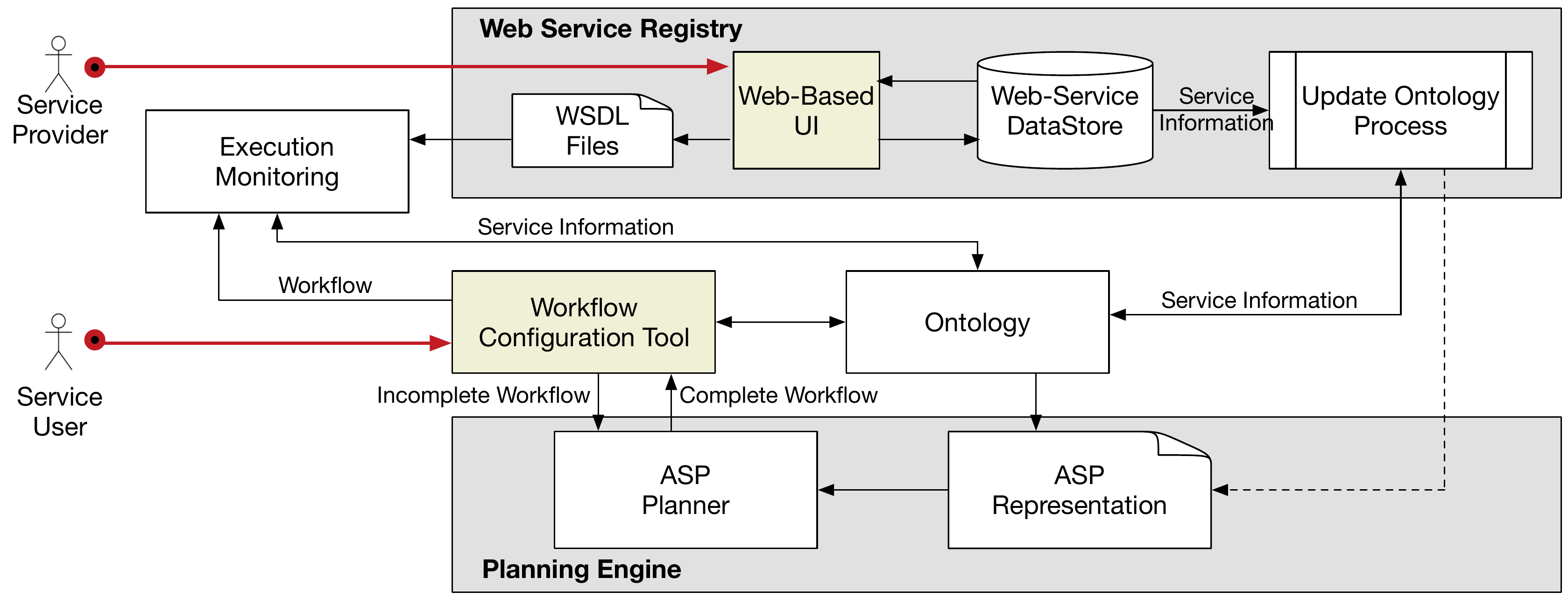}
  \caption{The Phylotastic Web Service Composition Framework}
  \label{overview}
\end{figure}

Figure~\ref{overview} shows an overview of our web service composition framework for Phylotastic. It consists of
a web service registry, an ontology, a planning engine,  a web service execution monitoring system, and a
workflow description tool. The flow of execution of the architecture starts with the \emph{workflow description tool}---a graphical
user interface that allows the user to provide information about the desired requirements of the phylogenetic trees generation
or extraction process.
The information collected from the user interface are mapped to the components of a planning problem instance, that will drive the
web service composition process. The planning problem instance representing the web service composition problem is obtained by
integrating the user goals with the description of web services, obtained from the service registry and the ontology. The planning engine
is responsible for deriving an executable workflow, which will be enacted and monitored by a web service execution system. The final outcome
of the service composition and execution is presented  to the user using the same workflow description tool.


Thus, the   components of the Phylotastic web service composition framework are:
\begin{list}{$\bullet$}{\itemsep=0pt \topsep=1pt \parsep=0pt \leftmargin=10pt} 

\item \emph{The Phylotastic Ontology}: this ontology is composed of
two parts: an ontology that describes the artifacts manipulated by the services (e.g., alignment matrices, phylogenetic trees, species names)  (the CDAO ontology \cite{cdao1}) and an 
ontology that describes the actual operations and transformations performed by the services.  Each class of services is associated with a name, inputs, parameters, and outputs. Instances of the services will also be associated with the data formats of their inputs, outputs, and parameters.  
For example, a service in the class \textit{taxon\_based\_ext} takes \textit{bio\_taxa} as an input and produces outputs \textit{species\_tree} and \textit{http\_code}. An instance of this class is \textit{get\_PhyloTree\_OT\_V1} whose input (\textit{bio\_taxa}) has the \textit{list\_of\_strings} format and its outputs (\textit{species\_tree}, \textit{http\_code}) have the \textit{newickTree} and \textit{integer} format, respectively. This information can be encoded in ASP as follows:\\
\texttt{\small
\hspace*{.5cm}op\_cl(tree\_ext). op\_cl(taxon\_based\_ext). op(get\_PhyloTree\_OT\_V1). \\
\hspace*{.5cm}subcl(taxon\_based\_ext,tree\_ext). cl(bio\_taxa). cl(species\_tree). \\
\hspace*{.5cm}t\_of(get\_PhyloTree\_OT\_V1,taxon\_based\_ext).\\
\hspace*{.5cm}has\_input(taxon\_based\_ext,set\_of\_names\_1,bio\_taxa). \\
\hspace*{.5cm}has\_output(taxon\_based\_ext,phylo\_tree\_1,species\_tree). \\
\hspace*{.5cm}has\_output(taxon\_based\_ext,http\_code\_1,http\_code).\\
\hspace*{.5cm}has\_inp\_df(get\_PhyloTree\_OT\_V1,bio\_taxa,list\_of\_strings). \\
\hspace*{.5cm}has\_out\_df(get\_PhyloTree\_OT\_V1,species\_tree,newickTree). \\
\hspace*{.5cm}has\_out\_df(get\_PhyloTree\_OT\_V1,http\_code,integer).
}

\item \emph{Web Service Composition as Planning}:  
In Phylotastic, we adopt the  view, advocated by several researchers, of mapping the web service
composition problem to a \emph{planning problem} \cite{compose4,mcisonzen01,compose7}. In this perspective, available web services
are viewed as \emph{actions} (or \emph{operations}) that can be performed by an agent, and the problem of determining the overall workflow can be reduced to a planning problem.
 In general, a planning problem can be described as a five-tuple $(S, S_0, G, A , I')$, where S is set of all possible states of the world, $S_0 \subseteq S$ denotes the initial state(s) of the world, $G \subseteq S$ denotes the goal states  the planning system attempts to reach, $A$ is the set of actions   to change one state of the world to another state, and the transition relation $I' \subseteq S \times A \times S$ defines the precondition and effects for the execution of each action. In term of web services, $S_0$ and $G$ are the initial state and the goal state specified in the requirement of web service requesters (e.g., the available input and the desired output of the workflow). The set  $A$ is a set of available services; $I'$  describes the effect of the execution of each service (e.g., data produced).
 
\item \emph{Web Service Engine}: The planning engine implemented in the Phylotastic project employs  
Answer Set Planning (ASP) \cite{Lifschitz02} and is responsible
for creating an executable workflow from the incomplete workflow and/or from the user specifications. The basic
planning algorithm has been discussed by \citeN{NguyenSP18}. This engine differs from the usual ASP-planning system 
in that it uses a two-stage process in computing solutions. In the first stage, planning is done at the 
abstract-level and the engine considers only service classes, matching their inputs and outputs. The result is 
a workflow whose elements are web services described at the abstract level. The second stage instantiates 
the abstract web services from the first stage with concrete services. In the process, it might need
to solve another planning problem, to address the issues of mismatches between formats of different 
services; for example,  there are several concrete services to recognize gene names and their outputs are sets
of scientific names of the genes; however, they are saved in different formats. 
This program consists of following ASP-rules encoding the operations and the initial state. 
The key rules are 

\begin{list}{$\circ$}{\itemsep=0pt \topsep=1pt \parsep=0pt} 
\item  {\texttt{\small ext(I,0) :- initially(I,DF$_I$).}} \:\: This rule states that the resource \texttt{I} with data format \texttt{DF$_I$} exists at the time moment 0. 

\item The rule encoding the operations and their executability are as follows: \\
\texttt{\small
\hspace*{.5cm} \{executable(A,T)\} :- op\_cl(A).   \\
\hspace*{.5cm} :- executable(A,T), has\_input(A,N,I), not match(A, I, T).  \\
\hspace*{.5cm} p\_m(A,I,T,O,T$_1$) :- op\_cl(A), has\_input(A,N,I), T$_1{\le}$T,  \\
\hspace*{4cm} ext(O,T$_1$),subcl(O,I).  \\
\hspace*{.5cm} 1 \{map(A,I,T,O,T$_1$) : p\_m(A,I,T,O,T$_1$)\}1.\\
\hspace*{.5cm} match(A,I,T) :- map(A,I,T,\_,\_). 
}
\item The next rules are used to generation operation occurrences:  \\
\texttt{\small
\hspace*{.5cm} 1 \{ occ(A,T): op\_cl(A) \} 1.   \\
\hspace*{.5cm} :- occ(A,T), not executable(A,T).\\
\hspace*{.5cm} ext(O,T+1) :- occ(A,T), has\_output(A,N,O). 
}
\end{list} 
In all the rules, \texttt{T} or \texttt{T1} denotes a time step.  
 From now on, we will denote with $\Pi_L$ the ASP-program for web service composition in the Phylotastic 
project.  

\end{list} 
\section{Selecting The Most Preferred Workflow: A Qualitative Approach}

Currently, $\Pi_L$ receives from the Workflow Configuration Tool a description of the desired workflow, e.g., 
desired input, desired output, specific classes of operations that should occur in the workflow.
Using
this information, it generates 
a concrete workflow meeting the desired requirements, 
sends it to the execution monitoring component which will execute the workflow and output 
the desired phylogenetic tree.  Frequently, there are several ways to construct a tree given the input, i.e., 
there are several solutions that $\Pi_L$ could return. Due to the differences in web services, 
not all solutions will produce the same result at execution (e.g., produce the same phylogeny), e.g., because of lack of agreement on the
evolutionary relationships between certain species or the ambiguity of certain species names. 
 In this paper, we present two enhancements of the 
system. In both enhancements, the notion of a \emph{most preferred workflow} is defined and 
users can interact with the system to select their most preferred workflow(s).

One way to compare the workflows  is to rely on the notion of  \emph{quality of service (QoS)} of web services.
QoS of a web service can be used as a discriminating factor that differentiates  
functionally similar web services. In general, QoS of a web service is characterized by several attributes, such as performance, reliability, scalability, accuracy, integrity, availability, and accessibility \cite{RajendranB09}. For 
the web services used in the Phylotastic project, we  collect the following attributes that influence the performance of a web service: {\bf (1)} \emph{response time}, {\bf (2)} \emph{throughput},
{\bf (3)} \emph{availability}, {\bf (4)} \emph{reliability}. Specifically, 
\begin{list}{$\bullet$}{\itemsep=0pt \parsep=1pt \topsep=1pt \leftmargin=10pt}
    \item {\em Response time}: Given a service $s$, the response time $q_{rt}(s)$ measures the  delay in 
    seconds between the moment when a request is sent and the moment when the results are received.
    \item {\em Throughput}: $q_{tp}(s)$ is the average number of successful responses for a give period of time.
    \item {\em Availability}: The availability $q_{av}(s)$ of a service $s$ is the probability that the service is accessible for a given period of time. 
    The value of the availability of a service $s$ is computed using the following expression $q_{av}(s) = T_a(s) / 
    \theta$, where $T_a$ is the total amount of time (in seconds) in which service $s$ is available during the 
    last $\theta$ seconds ($\theta$ is a constant set by an administrator of the service community).
        
    \item {\em Reliability}: The reliability $q_{re}(s)$ is the average operation time of service $s$ in which service $s$ is accessible and processes clients     requests successfully. It is measured by total operation time of  service $s$ divided by the number of failures. 
\end{list}
The quality vector of a service $s$ is denoted by the tuple  
$q(s) = (q_{rt}(s), q_{tp}(s), q_{av}(s), q_{re}(s))$. For the web services in the Phylotastic project, this information 
is maintained in the Service Registry (along with the ontology-based description of each service). 
We next define the QoS of a workflow based on the QoS of the web services. 

\subsection{QoS of Workflows}
\label{qos_composite}


Let $p = (s_1,s_2,...,s_n)$ be a workflow of web services. The quality of services of $p$, denoted by $q(p)$, 
is defined by  $q(p) = (q_{rt}(p), q_{tp}(p), q_{av}(p), q_{re}(p))$ where 
\begin{list}{$\bullet$}{\itemsep=0pt \parsep=0pt \topsep=0pt \leftmargin=10pt}
    \item {\em Response time}: The response time $q_{rt}(p)$ is defined as total response time of all  services in the workflow $p$: $q_{rt}(p) = \sum_{i=1}^{n} q_{rt}(s_i)$. 
    
    \item {\em Throughput}: The throughput $q_{tp}(p)$ of plan $p$ is the average of the throughputs of the services that participate in $p$: $q_{tp}(p) = \frac{\sum_{i=1}^{n} q_{tp}(s_i)}{n}$. 
    
    \item {\em Availability}: In general, the availability of the services for $p$ should be defined as 
    $q_{av}(p) = \prod_{i=1}^{n} Pr(s_i \mid s_1,\ldots,s_{i-1})$ where $Pr(s_i \mid s_1,\ldots,s_{i-1})$ is the conditional probability of $s_i$ is available given that $s_1,\ldots,s_{i-1}$ have been successfully executed. For simplicity of  representation, we assume that the services are mutually independent, then  $q_{av}(p) = \prod_{i=1}^{n} q_{av}(s_i)$. In our current implementation, we use $q_{av}(p) = \min \{q_{av}(s_i) \mid i =1,\ldots,n\}$, as an approximation which avoids extensive floating point operations.

%
%
    \item {\em Reliability}: The reliability $q_{re}(p)$ is calculated as the average of reliability values of element services in $p$: $q_{re}(p) = \frac{\sum_{i=1}^{n} q_{re}(s_i)}{n}$. 
\end{list}

 \subsection{ASP Encoding of QoS}

The QoS of a workflow can be computed using ASP as follows. 
As with the actions used in the web composition process, we extract the QoS information of services and represent it as ASP-facts  
of the forms: $has\_qos\_rt(s, v),has\_qos\_av(s, v),has\_qos\_tp(s, v),has\_qos\_re(s, v)$ where $s$ is the service identifier, and $v$ is 
the QoS value of the corresponding attribute. The computations of QoS attributes for a workflow are encoded as following:

%
%
%

{\footnotesize
\[
	\begin{array}{ll}
	qos\_rt\_wf(RT_{W}) & {:}{- }\quad RT_{W} = \#sum\{RT,T,X : occ\_concrete(X,T),has\_qos\_rt(X,RT)\}. \\
	qos\_tp\_wf(TP_{S}/nsteps)& {:}{-}\quad TP_{S} = \#sum\{TP,T,X : occ\_concrete(X,T),has\_qos\_tp(X,TP)\}.\\
	qos\_re\_wf(RE_{S}/nsteps) &{:}{-}\quad RE_{S} = \#sum\{RE,T,X : occ\_concrete(X,T),has\_qos\_re(X,RE)\}.\\
          qos\_av\_wf(AV_{W}) &{:}{-}\quad AV_{W} = \#min\{AV,T,X : occ\_concrete(X,T),has\_qos\_av(X,AV)\}.
\end{array}
\]}

In the above rules, $nsteps$ is the number of services  in the plan
that is computed by the planning module. 
Since the QoS of a service (or a workflow) is a tuple of values representing different attributes, 
there are different ways for comparing services (or workflows). 
Different users might have different preferences over these attributes (e.g., response time is the most important factor, or reliability is the most important factor, etc.). 
We discuss two possibilities: 

\begin{list}{$\bullet$}{\itemsep=0pt \topsep=1pt \parsep=1pt \leftmargin=10pt} 
\item \emph{Weighted QoS}: A user specifies the weights $W_{rt}$, $W_{tp}$, $W_{av}$, and $W_{re}$ that he/she would like to assign for 
the response time, the throughput, the availability, and the reliability, respectively. 
The weighted QoS of a plan $p$ is then computed by $w(p) = q_{rt}(p)*W_{rt} +q_{tp}(p)*W_{tp} + q_{av}(p)*W_{av} + q_{re}(p)*W_{re}$. 
Under this view, $w(p)$ can be computed as follows: 
$
 \hspace*{.5cm}score\_qos\_wf(Sc)\:{:}{-} \:\:\:\: qos\_rt\_wf(RT_{W}),wei\_rt(W_{rt}),qos\_tp\_wf(TP_{W}),wei\_tp(W_{tp}), \newline  
 \hspace*{3.5cm}qos\_re\_wf(RE_{W}),wei\_re(W_{re}),qos\_av\_wf(AV_{W}),wei\_av(W_{av}),  \newline
 \hspace*{3.5cm}Sc = RT_{W}*W_{rt} + TP_{W}*W_{tp} + RE_{W}*W_{re} + AV_{W}*W_{av}.
$

\noindent To select workflows with the best QoS, we will only need to add the statement 
\[
\mathit{\#maximize\{ScoreQoS : score\_qos\_wf(ScoreQoS)\}.} 
\]

\item \emph{Specified Preferences QoS}: An alternative to the weighted QoS is to allow users to specify a partial ordering over the set of attributes
that will be used in identifying most preferred workflows by a lexical ordering in accordance to the preferences. 
For example, assume that the preference ordering is $x_1 > x_2 > x_3 > x_4$ where $x_i > x_j$ means that attribute $x_i$ is 
 preferred to the 
attribute $x_i \ne x_j$ and  $x_i  \in \{rt, av, tp, re\}$. 
As the values in the QoS of a service behave differently, we write $q_{x}(s) \prec q_{x}(s')$ to denote that $s$ 
is better than $s'$ w.r.t. the attribute $x$.  
 The most preferred workflow is defined via a lexicographic ordering: $p \prec p'$ if there is $1 \leq i \leq 4$ such that $q_{x_j}(p) = q_{x_j}(p')$ for $j < i$ and $q_{x_i}(p) \prec q_{x_i}(p')$. 
 This can easily be implemented using the priority level in \textbf{clingo}. \\
\textit{\small 
  \#maximize \{$q_{x_1}$@4\}.\:\:\:\#maximize \{$q_{x_2}$@3\}.\:\:\:\#maximize \{$q_{x_3}$@2\}.\:\:\:\#maximize \{$q_{x_4}$@1\}.
} 

\noindent For a concrete example, assume that the preference ordering is $rt > re > tp > av$, the corresponding ASP encoding is as follow:
\begin{align*} 
  \#maximize \{RT_{W}@4 : qos\_rt\_wf(RT_{W})\}. \quad
  \#maximize \{RE_{W}@3 : qos\_re\_wf(RE_{W})\}. \\
  \#maximize \{TP_{W}@2 : qos\_tp\_wf(TP_{W})\}. \quad
  \#maximize \{AV_{W}@1 : qos\_av\_wf(AV_{W})\}. 
\end{align*}
\end{list} 
The above feature is  implemented in the Phylotastic system. 
In either case, we can display a set of workflows for the users to decide which workflow should be executed.

\section{Refining a Workflow}
\label{sub:replanning_problem}

Evidence that emerged in the Phylotastic project as described by \citeN{phylotastic1} shows that evolutionary biologists tend to develop their analysis protocols in an incremental manner, through successive refinements, often driven by the specific properties of the dataset being processed and the opportunities revealed by intermediate results.
 Users of the Phylotastic system also often have strong preferences about certain type of services that they want (or do not want) to use. For this reason,  the Workflow Configuration Tool  allows a user to update a given workflow and resubmit it to the ASP-planner. Presently, the ASP-planner considers this as a new request and restarts the computation. This approach is simple but also has some drawbacks. First, the new workflow and the original workflow can be very different in the services that they use, which is often unexpected to the user (and undesired, since changing services might lead to a different phylogeny). Second, this approach can be computational expensive as it cannot reuse the original workflow. We propose an approach to 
 address the changes requested by a user that can preserve \emph{as much as possible the original workflow}. We focus on the following four categories of changes:
\begin{list}{$\bullet$}{\itemsep=0pt \parsep=1pt \topsep=1pt}
    \item {\em IO request}: Request to change input and/or output.
    \item {\em Avoidance request}: Avoid using one class of services.
    \item {\em Inclusion request}: Request that a particular service to be used in the workflow.
    \item {\em Insertion request}: Request that a service is inserted at a particular position.   
\end{list}

\subsection{Similarity Between Workflows: Formalization} 
\label{sub:similarity_lib}
{ 

Given two workflow $p$ and $p'$, we define the concept of  \emph{similarity} between $p$ to $p'$.  
In the following, we view a workflow as a directed acyclic graph  $G = (V, E)$, where  $V$ is the set of nodes  and each node is a service; $E$ is the set of edges and each edge is an exchange of resources between two services in the workflow. Observe that each service $v$ will have a set of inputs, a set of outputs, and some description. 
For each service $v$, $input(v)$ and $output(v)$ denote the set of inputs and outputs of $v$, respectively.  
The \emph{similarity between two workflows} is defined as a combination of \emph{nodes similarity}, \emph{edges similarity}, and \emph{contextual and topology similarity} \cite{BeckerL12,AntunesBBCDFDFG15}. 
Let $G_1 = (V_1, E_1)$  and $G_2 = (V_2, E_2)$ be two graphs. 
The similarity between $G_1$ and $G_2$, 
denoted by  \texttt{sim\_workflows($G_1,G_2$)},  is defined next.  

\begin{list}{$\bullet$}{\itemsep=0pt \parsep=1pt \topsep=1pt \leftmargin=10pt}
\item 
{\bf Node Similarity.} Let $v_{1} \in V_1$ and $v_{2} \in V_2$.  We fist define
the similarity between two nodes and then use this measure to define the similarity between nodes of workflows. 
As a node represent a web service, the similarity of two nodes can be determined based on their mutual 
position in the ontology (note that the Phylotastic ontology classifies services based on a taxonomy of 
classes of services). Thus, two nodes can be considered to be  similar if they are close in the ontology, share the same 
inputs, outputs, or have similar descriptions. These features are considered in the following definitions. 
%
%
\begin{list}{$\circ$}{\itemsep=0pt \topsep=0pt \parsep=0pt \leftmargin=6pt}
	\item \texttt{sim\_nodes\_onto($v_{1},v_{2}$)}: measures the similarity between $v_{1}$ and $v_{2}$ by considering them as nodes in the ontology: \\
			$sim\_nodes\_onto(v_{1},v_{2}) = \frac{1}{1 + d\_nodes\_onto(v_{1},v_{2})}, \:\: \textnormal{where}$\\
		   $d\_nodes\_onto(v_{1},v_{2}) = path\_len(lca(v_{1},v_{2}),v_{1}) + path\_len(lca(v_{1},v_{2}),v_{2})$.
		   Intuitively, the similarity between two nodes in an ontology 
		   is disproportional to the distance ($path\_len(.)$) between their lowest common ancestors ($lca(.)$) of 
		   them.

	
	\item \texttt{sim\_nodes\_inp($v_{1},v_{2}$)}: measures the similarity the nodes by considering their inputs, the more inputs they share the more similiar they are. Thus, \\
			$$sim\_nodes\_inp(v_{1},v_{2}) = 2*\frac{| input(v_{1}) \cap input(v_{2}) |}{| input(v_{1}) | + | input(v_{2}) |}.$$ 
	
	\item \texttt{sim\_nodes\_oup($v_{1},v_{2}$)}: measures the similarity between two nodes by considering their outputs, computed similarly to \texttt{sim\_nodes\_oup($v_{1},v_{2}$)}. So,\\
			$$sim\_nodes\_oup(v_{1},v_{2}) = 2*\frac{| ouput(v_{1}) \cap output(v_{2}) |}{| output(v_{1}) | + | output(v_{2}) |}.$$
	
	\item \texttt{sim\_nodes\_des($v_{1},v_{2}$)}: measures the similarity between the English descriptions of the two nodes. 
	We use off-the-shelf libraries {\small \tt Stanford CoreNLP} \footnote{ {\small \url{https://stanfordnlp.github.io/CoreNLP}}}, 
	{\small \tt NLTK}\footnote{ {\small \url{http://www.nltk.org/}}}, and {\small \tt Scikit-Learn}\footnote{ \small{\url{http://scikit-learn.org/}}} to 
	process the descriptions and transform these text descriptions to a matrix of {\small \tt TF-IDF} (term frequency-inverse document frequency) of 
	the two documents. From each document we derive a real-value {\small \tt TF-IDF} vector;  the similarity index between two text 
	descriptions is computed based on cosine similarity between their {\small \tt TF-IDF} vectors.

\end{list}
The similarity between two nodes, denoted by \textbf{sim\_nodes($v_{1},v_{2}$)}, is then defined as the weighted sum between their four similarity 
measures 	  
%
			\[\begin{array}{ll} 
			sim\_nodes(v_{1},v_{2}) = &
			                                              w_{onto}*sim\_nodes\_onto(v_{1},v_{2}) +   
			                                              w_{inp}*sim\_nodes\_inp(v_{1},v_{2}) + \\
			                                             & w_{oup}*sim\_nodes\_oup(v_{1},v_{2}) + 
			                                              w_{des}*sim\_nodes\_des(v_{1},v_{2}) 
			                                              \end{array} 
			                                              \]
where $ w_{onto}, w_{inp}, w_{oup}, w_{des}$ are weight values assigned to each attribute, 
$w_x \in [0,1]$ ($x \in \{onto, inp, oup, des\}$) and $w_{onto} +w_ {inp} + w_{oup} + w_{des}= 1$. In our 
implementation, the values of $w_{onto}, w_ {inp} , w_{oup}, w_{des}$ are $0.6, 0.15, 0.15$ and $0.1$ 
respectively. The intuition behind these values lies in the fact that within an ontology, the similarity between objects depends heavily on their relative position to their lowest common ancestor; thus $w_{onto}$ should play the deciding factor. $ w_ {inp} = w_{oup}$ because of the symmetry between inputs and outputs. 
$w_{des}$ is smaller than $ w_ {inp}$ or $w_{oup}$ because our current ontology has different levels of detail in the 
English description of services.
%


%
Finally, \textbf{sim\_nodes\_workflows($G_1,G_2$)} is defined as follows:
		   \begin{equation} \label{simnode}
		   sim\_nodes\_workflows(G_1,G_2) = 2 * \frac{\sum_{v_1 \in V_1} \sum_{v_2 \in V_2} sim\_nodes(v_{1}, v_{2})}{|V_1| + |V_2|}\end{equation}

\item {\bf Edge Similarity.} Let $e_1 \in E_1$ and $e_2 \in E_2$. 
As an edge connected two services (nodes) in a workflow and denotes an exchange between two nodes (output of one is input of the other). 
For this reason, the similarity between two edges can be defined via the similarity between the nodes relating to them and the descriptions of their 
inputs and outputs. For an edge $e$, let $s(e)$ and $d(e)$ denote the source and destination of $e$, respectively; 
Furthermore, let $lab(e) = (o_{s(e)},i_{d(e)})$, where $o_{s(e)}$ is the
output of $s(e)$ that is used as the input $i_{d(e)}$ of $d(e)$. We define:

\begin{list}{$\circ$}{\itemsep=0pt \topsep=1pt \parsep=1pt \leftmargin=10pt}
	\item \texttt{sim\_ed\_nod($e_{1},e_{2}$)}: measures the similarity of two edges by considering the similarity of the nodes related to the edges and is defined by \\
			$sim\_ed\_nod(e_{1},e_{2}) = \frac{1}{2}*(sim\_nodes(s(e_1),s(e_2)) + sim\_nodes(d(e_{1}), d(e_{2})))$.
 
	\item \texttt{sim\_ed\_re($e_1,e_2$)}:  measures the similarity of two edges by considering distance between their labels
	and is defined by \\
			$sim\_ed\_re(e_{1},e_{2}) = \frac{1}{1 + d\_ed\_ont(lab(e_{1}),lab(e_{2}))}, \:\: \textnormal{where}$ 
	            $
                      d\_ed\_ont(lab(e_{1}),lab(e_{2})) = \frac{1}{2} *(d\_nodes\_onto(o_{s(e_1)} ,o_{s(e_2)}) + d\_nodes\_onto(i_{d(e_1)},i_{d(e_2)}))                                                                    
                      $.

\end{list}
The similarity between two edges, denoted by \textbf{sim\_edges($e_{1},e_{2}$)}, is then defined as the weighted sum between their two similarity 
measures.
$sim\_edges(e_1,e_2) =  w_{node} * sim\_ed\_nod(e_{1},e_{2})  + w_{label}*sim\_ed\_re(e_1,e_2)$. 
where $0 \le w_{node}, w_{label}\le 1$ are weight values of corresponding attributes contributions such that 
$w_{node} + w_{label} = 1$. In our implementation, we assign the values of $w_{node}, w_{label}$ are $0.5$ and $0.5$ respectively.
We define $$sim\_edges\_workflows(G_1,G_2) = 2 * \frac{\sum_{e_1\in E_1} \sum_{e_2 \in E_2} sim\_edges(e_{1},e_{2})}{|E_1|+|E_2|}.$$

\item {\bf  Topological Similarity.} 
By considering workflows as graphs, we can also consider their similarity based on the amount of changes needed to convert one to the other. The notion 
of an \emph{Edit-Distance}, denoted by $dist\_topo(G_1,G_2)$, between two graphs---the smallest number of changes (insertions, deletions, 
substitutions, etc.) required to transform one structure to another---has been introduced and algorithms for computing it have been 
developed by \citeN{ZhangS89}. The topological similarity between two graphs is defined by  
          $sim\_topo(G_1,G_2) =  \frac{1}{1 + dist\_topo(G_1,G_2)}$.

\end{list}

Having defined various types of similarities between elements of the workflows, we can now define the similarity between two workflows as a weighted sum of these similarities:  
  \[\begin{array}{ll} 
  sim\_workflows(G_1,G_2) =  & w_{no}*sim\_nodes\_workflows(G_1,G_2) +  \\
   & w_{ed}*sim\_edges\_wf(G_1,G_2) + w_{to}*sim\_topo(G_1,G_2)
  \end{array} \]
 where  $w_{no}, w_{ed}, w_{to} \in [0,1]$ and  
 $w_{no} + w_{ed} + w_{to} = 1$. 
  In our implementation, we use  $w_{no}=0.45, w_{ed}=0.35$  and $w_{to}=0.2$. Here, the emphasis is still the similarity between nodes. This is because the nodes are the main components of a workflow. 
  For this reason, we place greater emphasis  on the edges than the topology of a workflow, because the labels of the edges are also critical to the workflow. 
Due to the fact that the computation of the similarity between two workflows is deterministic,  deals mostly with real numbers,
 and the fact that new answer set solvers allow for the integration of external atoms as described by \cite{KaminskiSW17}, 
we implemented a package for computing the similarity between two workflows as a Python library and used this as external predicates in ASP.

\subsection{ASP-Code for Replanning}
\label{sub:asp_preferences_desires}

Given a workflow and some modifications requested by a user, the similarity measure introduced in the previous subsection can be used to select the 
workflow that satisfies the user's requests and is most similar to the original one. 
We next present the ASP implementation addressing each of the change categories discussed at the beginning of this section and then selecting the 
\emph{most similar workflow}. 
We will use a simple original workflow for generating a \emph{gene-species reconciliation tree} from \emph{a set of gene names} (Figure~\ref{original_workflow}) as a running example. In this workflow,  each node (circle) represents a service (e.g., $a$ is a service) and each connection between two nodes represent an edge with its label 
(e.g., the edge from $a$ to $b$ has $oa$ as an output of $a$ which is used as input $ib$ of $b$). Here,
$a$, $b$, $c$, $d$, $e$ and $f$ are the short name for the services  \emph{Get\_GeneTree\_from\_Genes, Ext\_Species\_from\_GeneTree, Resolved\_Names\_OT, Get\_PhyloTree\_OT\_V1, GeneTree\_Scaling\_V1} and 
\emph{Get\_ReconciliationTree} respectively. 
\begin{figure*}[h]
		\centerline{\includegraphics[width=0.65\textwidth]{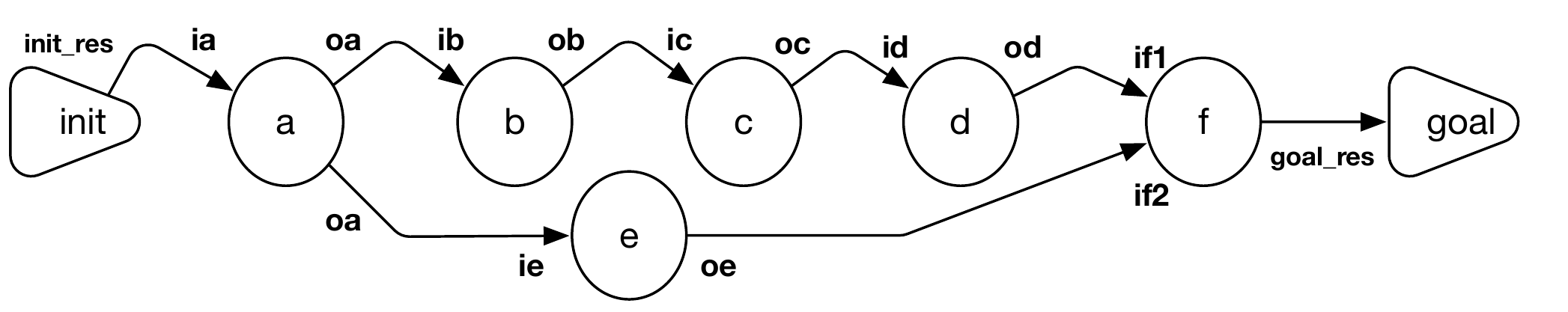}}
	\caption{Original workflow}
	\label{original_workflow}
\end{figure*}

Intuitively, the workflow in Figure~\ref{original_workflow} represents a workflow generated by the planning engine, encoded   by the set of atoms: 
\begin{align*} 
 \{initially(init\_res,dfi),occ(a,1), occ(b,2),occ(c,3), occ(d,4), 
  occ(e,5),occ(f,6), \\ finally(goal\_res,dfg), map(a,ia,1,init\_res,0), 
 map(b,ib,2,oa,2), map(c,ic,3,ob,3), \\ 
 map(d,id,4,oc,4), 
 map(e,ie,5,oa,2), map(f,if1,6,od,5), map(f,if2,6,oe,6).\}
\end{align*} 
In this encoding, $initially/2$ ($finally/2$) states that the input (goal) with its data type; $occ(x,i)$ states that the service $x$ must occur at the step 
$i$; $map(s, i, t1, o, t2)$ says that an output $o$ of step $t2$ is an input $i$ of service $s$ at step $t1$.
\subsubsection{IO Request}


For an IO request, all that needs to be changed is the ASP encoding sent to the ASP-planner, specifically atoms of the form \texttt{\small initially} and/or \texttt{\small finally} will be updated to reflect the request. The planning engine will be executed and returned the most similar workflow to the current one.  



\subsubsection{Avoidance Request}

A request to avoid using a service $s$ (or a class of services $c$) will be translated to the atom $do\_not\_use(s)$ 
($do\_not\_use(c)$) and supplied to the planning engine. 
We use the following ASP-rules to enforce this request:  
\begin{align}
  is\_used(C)\: {:}{-} \:\:\:\: member(X, C), \: occ(X, \_). \label{r1}\\
  is\_used(X)\: {:}{-} \:\:\:\: occ(X, \_). \label{r3}\\  
  {:}{-} \:\:\:\: is\_used(X), \: do\_not\_use(X). \label{r3}
\end{align}
The first two rules determine the class of service \texttt{C} or the service $X$ is used in the workflow. 
The third enforces the request of the user to avoid the use of the service or a class of services.  
Some possible resulting workflows satisfying the request $do\_not\_use(d)$ are shown in Figure \ref{avoid_service_wf}.
%
%
\begin{figure*}[t]
		\centerline{\includegraphics[width=0.95\textwidth]{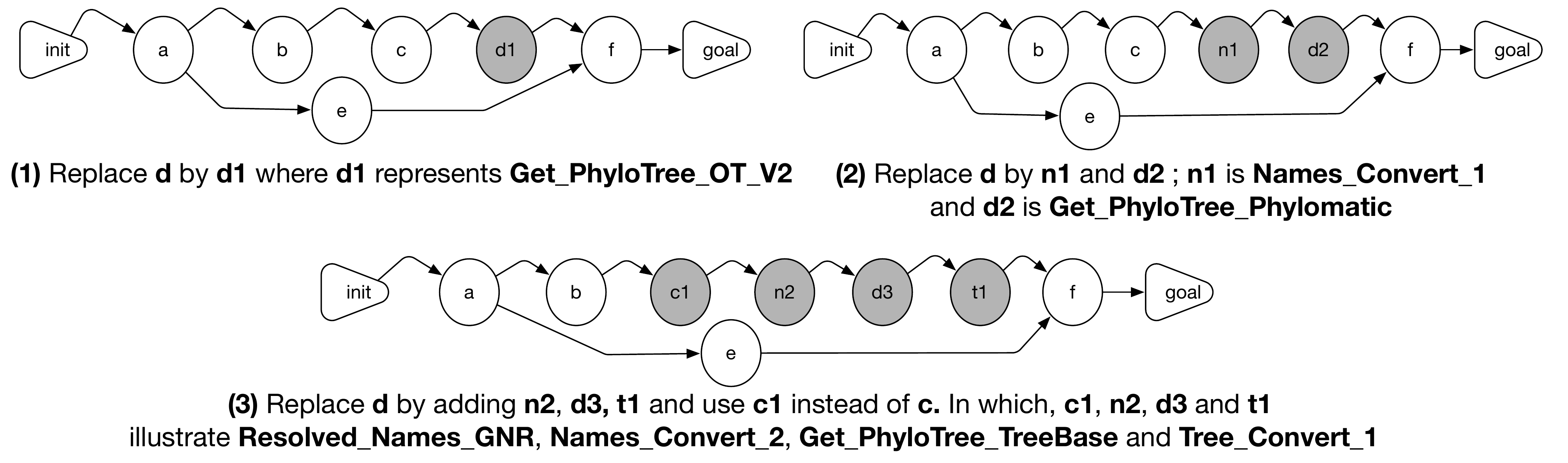}}
	\caption{Some updated workflows with avoidance the service $d$}
	\label{avoid_service_wf}
\end{figure*}
\subsubsection{Inclusion Request} 
A request to include a service $s$ (or a class of services $c$) will be translated to the atom $must\_used(s)$ 
($used(c)$) and supplied to the planning engine. The rules \eqref{r1}-\eqref{r3}  are used with  
the rule:  
%
%
\begin{align} 
  {:}{-} \:\:\:\: must\_used(X), \: not \:\: is\_used(X).  \label{r4}
\end{align} 
to make sure that the request to include some service is satisfied.  
Figure \ref{include_service_wf} displays some possible updated workflows with the request of using \textbf{(1)} \texttt{\small e2} (\emph{GeneTree\_Scaling\_V2}) 
or  \textbf{(2)} \texttt{\small e3} (\emph{GeneTree\_Scaling\_V3})
\begin{figure*}[h]
		\centerline{\includegraphics[width=1.0\textwidth]{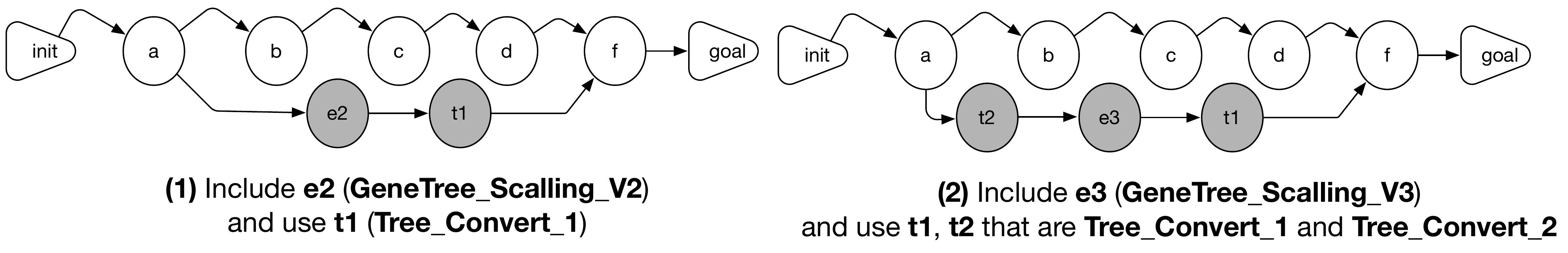}}
	\caption{Updated workflows with inclusion request of \texttt{\small e2} or \texttt{\small e3}}
	\label{include_service_wf}
\end{figure*}
\subsubsection{Insertion Request} 
This is a special case of an inclusion request. Specifically, it specifies where the service should be included. This request is translated into the ASP atoms of the form $before(x,y)$ (service $x$ must be executed before service $y$) and $must\_used(x)$. To implement this, we add to the 
%
before but also some after statements}
\begin{align}
  is\_before(C,D)\: {:}{-} \:\:\:\:  occ(C,T), \: occ(D,T_1), \: T<T_1. \label{r5} \\
  {:}{-} \:\:\:\: before(C,D), \: not \: \: is\_before(C,D). \label{r6} 
\end{align} 
Figure \ref{ordered_serives_wf} shows some workflows accommodating the request ``\emph{use service $e$ after $a$ and before $b$}'', encoded by
$ \{must\_used(e), \:  before(a,e), \: before(e,c).\}$.   
%
\begin{figure*}[h]
		\centerline{\includegraphics[width=1.0\textwidth]{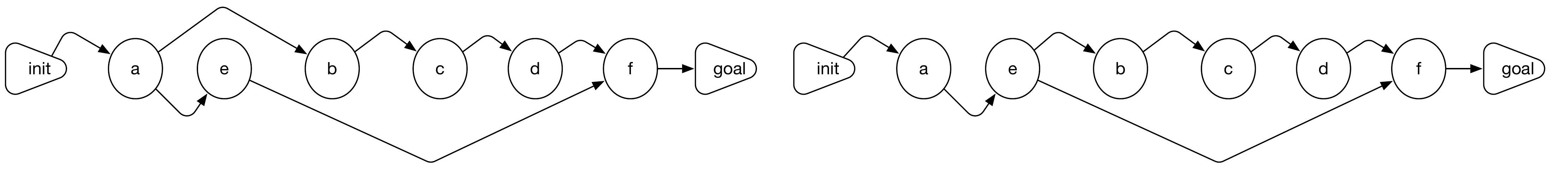}}
	\caption{Service e is executed after a and before c}
	\label{ordered_serives_wf}
\end{figure*}
\subsubsection{Selecting a Most Similar Workflow}

Let $\Pi_R$ be the set of rules \eqref{r1}-\eqref{r6} for enforcing the requests of the users and $C$ denote the set of atoms encoding the requests of an user. Our goal is to compute a new workflow that satisfies $C$ and is as  similar as possible to the original workflow. It is easy to see that $\Pi_L \cup \Pi_R \cup C$ will return workflows satisfying $C$. As such, we only need to identify, among all solutions provided by $\Pi_L \cup \Pi_R \cup C$, the workflow that is most similar to the original workflow. This can be achieved by encoding the original workflow, providing it as input, and exploiting the Python package for computing the similarity of two workflows (subsection~\ref{sub:similarity_lib}). 
To do so, let us assume that the original workflow is encoded  as a set of atoms $O$ of the form $old\_occ(s,t)$ and 
$old\_map(s, i, t_1, o, t_2)$. There are different possibilities here: 

\begin{list}{$\bullet$}{\itemsep = 0pt \parsep=1pt \topsep=1pt \leftmargin=10pt} 
\item {\bf Computing exact value of similarity using ASP}: theoretically, the exact value of similarity between the constructed workflow and the original one can be computed in the ASP using an external call to $sim\_workflows$ and the most similar workflow can be computed using the 
$\mathtt{maximize}$ statement by 
\begin{align} 
sim(V)\: {:}{-}\:\:\:\: V = @sim\_workflows(wf_1, wf_2). \\
\# \mathtt{maximize} \{V : sim(V)\}. 
\end{align}
where $wf_1$ and $wf_2$ are two sets of atoms encoding the two workflows (the original and the computed one), this is essentially the sets 
of atoms of the forms $old\_occ(.)$, $old\_map(.)$, $occ()$, and $map()$ that is generated by the planning engine and supplied in $O$.  
This means that we need a set data structure to implement this approach. We did not implement this due to the fact that  
{\bf clingo} does not yet provide a construct for set. We hope to work with the {\bf clingo} group to introduce set as a basic data type for use besides the aggregate functions. 

%

\item {\bf Approximating the value of similarity using ASP}: Since the set of facts of the form $occ(s,t)$ correspond to the nodes in the workflow, we can approximate the similarity of  two workflows by considering only the similarity between nodes of the two workflows (old and new). This can be realized by the following rules: 
\begin{align*} 
single\_sim\_nodes(X,Y,Z)\:{:}{-}\:\:\:\: old\_occ(X,I_1),occ(Y,I_2), Z=@sim\_nodes(X,I_1,Y,I_2).\\
sum\_sim\_nodes(S)\: {:}{-}\:\:\:\: S = \#sum\{Z,X,Y : single\_sim\_nodes(X,Y,Z)\}. \\
sim\_nodes\_workflows(R)\: {:}{-}\:\:\:\: sum\_sim\_nodes(S), NO =  \#count \{X,I_1: old\_occ(X,I_1)\}, \\
 NN = \#count\{Y,I_2: occ(Y,I_2)\}, R = 2*S/(NO + NN). \\
 \#\mathtt{maximize}\{R : sim\_nodes\_workflows(R)\}. 
\end{align*} 
It is easy to see that the above rules implement formula \eqref{simnode}.  

\item {\bf Computing exact value of similarity using multi-shot ASP}: 
This approach has been implemented using the multi-shot ASP \cite{KaminskiSW17}. Basically, a Python wrapper is used to control the search for the most similar workflow to the original one. It implements the following loop: \\
\texttt{\footnotesize
\hspace*{.25cm}  \textbf{for} each answer set of $\Pi_L \cup \Pi_R \cup C$ \\ 
\hspace*{.5cm} compute the similarity $v$ of the solution and the original workflow \\
\hspace*{.5cm}  \textbf{if} it is greater than the current value (initiated with -1) \\
\hspace*{.5cm} \textbf{then} keep the solution 
}

\end{list} 
\subsection{Refining a Workflow: Case-Studies}
We illustrate the new features of our system using three use cases developed in the Phylotastic project. 

\subsubsection{From a Set of Gene Names to a Reconciliation Tree in \texttt{\small newick} Format} 
\begin{figure}[h]
	\centering	\includegraphics[width=\textwidth]{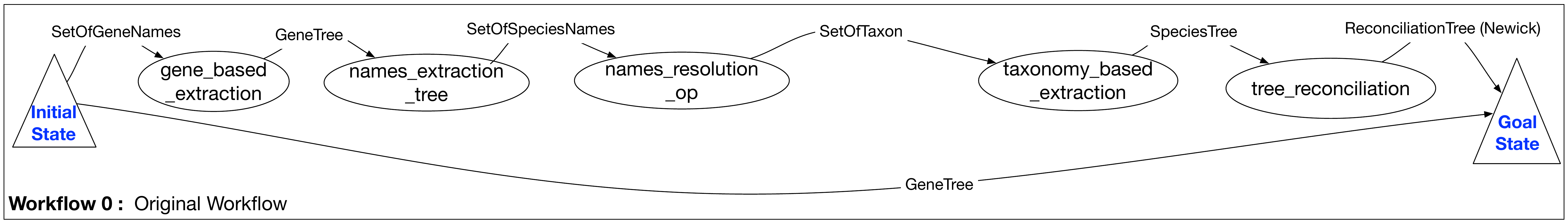}
	\caption{Original Workflow \label{eval_original_workflow}}
\end{figure}

In this use case, the user wants to generate a 
phylogenetic reconciliation tree in the \texttt{\small newick} format from a set of gene names, whose format is 
\texttt{\small set\_of\_strings}. 
The user can use the  {\em Workflow Configuration Module} to design an initial workflow with two nodes (initial node and goal node) 
with this information as described by \citeN{NguyenSP18}. The module will then convert this information into ASP-atoms \texttt{\small initially(setOfGeneNames,set\_of\_strings)} and \texttt{\small finally(reconciliationTree,newickTree)}, which will be sent 
the planning engine.  
The result is the workflow shown in Figure \ref{eval_original_workflow}, where the triangles identify the input and output and the name of the service that should be executed at each step. For example, a \texttt{\small gene\_based\_extraction} service should be executed at step 1 and a \texttt{\small names\_extraction\_tree} service needs to be executed at step 2. The workflow shown in the figure is also the one with highest QoS (1.4224).

The user, after examining  the workflow,  determines that the input \texttt{\small GeneTree} of \texttt{\small tree\_reconciliation} service needs to be scaled before being processed by \texttt{\small tree\_reconciliation}; 
and, this requires that the service \texttt{\small gene\_tree\_scaling} should be inserted after \texttt{\small gene\_based\_extraction} and before  \texttt{\small tree\_reconciliation} service. 
The experimentation has been performed on a machine running MacOS 10.13.3 with 8GB DDRam 3 and a 2.5GHZ Intel-Core i5 (3rd Generation)
with 54 classes of services and 125 concrete specific services in Ontology domain. 

\begin{list}{$\bullet$}{\itemsep = 0pt \parsep=1pt \topsep=1pt \leftmargin=10pt}  

\item {\bf Approximating the value of similarity using ASP}: Using this approach, there are different updates to the original workflow. 
Four of them with the highest similarity values are displayed in Figure \ref{eval_original_workflow_app_2_3}. The highest similarity value of 0.5779 comes 
from  \texttt{\small$WF_2$} and \texttt{\small$WF_3$}. Observe that these workflows have only one modification (\texttt{\small gene\_tree\_scaling} 
occurs at step 4 in \texttt{\small$WF_2$} and step 1 in \texttt{\small$WF_3$}). Furthermore, \texttt{\small$WF_1$} has two changes 
(adding \texttt{\small gene\_tree\_scaling} and replacing \texttt{\small taxonomy\_based\_extraction}  by \texttt{\small phylogeny\_based\_extraction}); 
and \texttt{\small$WF_4$} has three changes. The total processing time of this approach is $35.183$ seconds.

\item {\bf Computing exact value of similarity using multi-shot ASP}: The previous approach only approximates the similarity values of the new solutions 
and the original workflow. Using the multi-shot ASP, we can calculate the exact value of the similarity. Figure \ref{eval_original_workflow_app_2_3} shows 
the same 4 workflows considered in the previous approach with their exact value of similarity. The clear winner is $WF_2$. The 
system takes $37.937$ seconds to compute these updates.
\begin{figure}[t]
	\centering	\includegraphics[width=\textwidth]{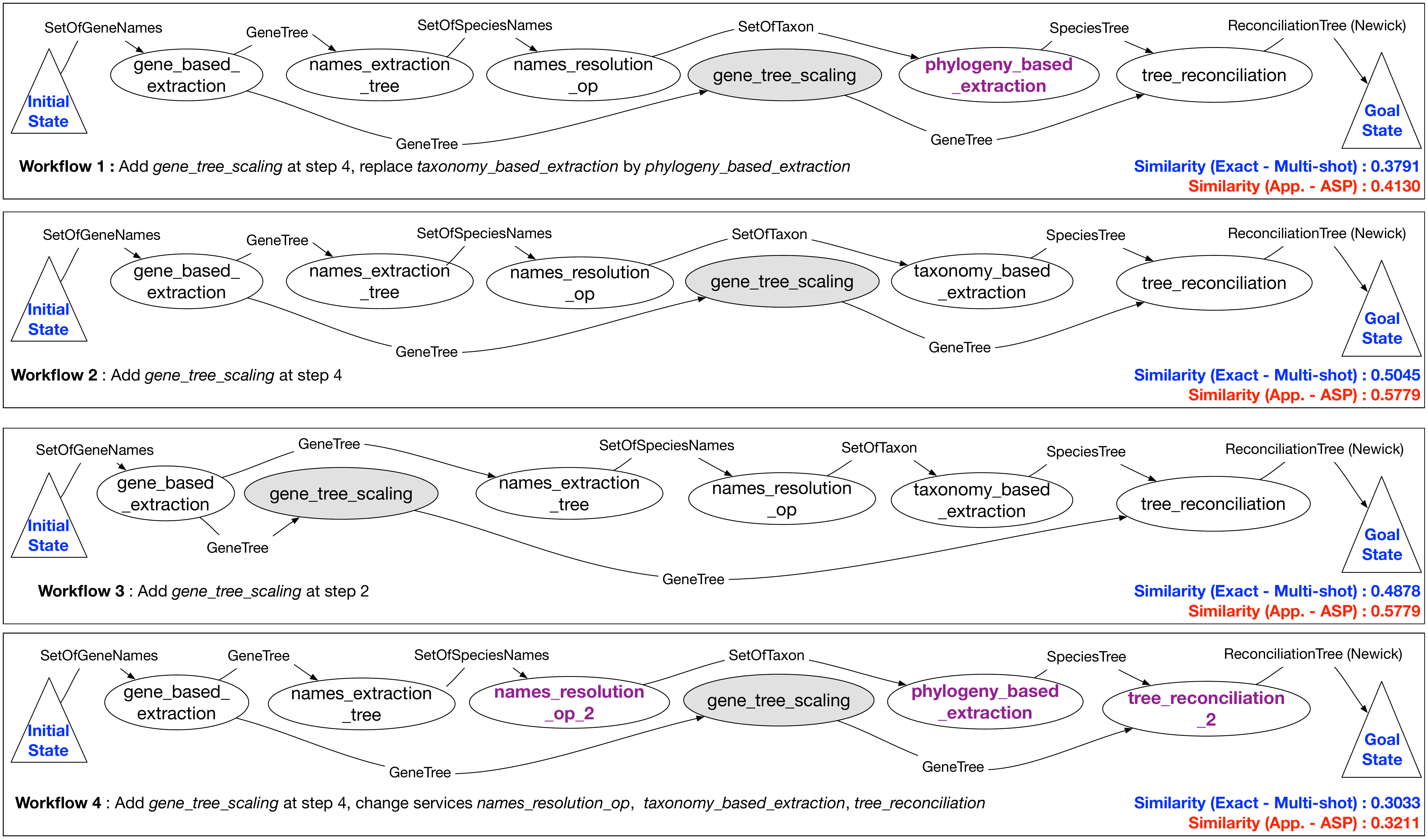}
	\caption{Possible updated workflows by 2 approaches  \label{eval_original_workflow_app_2_3}}
\end{figure}
\end{list}

\subsubsection{Generating a Species Tree in  \texttt{\small newick} Format from Free Plain-Text}

\begin{figure}[t]
	\centering	\includegraphics[width=\textwidth]{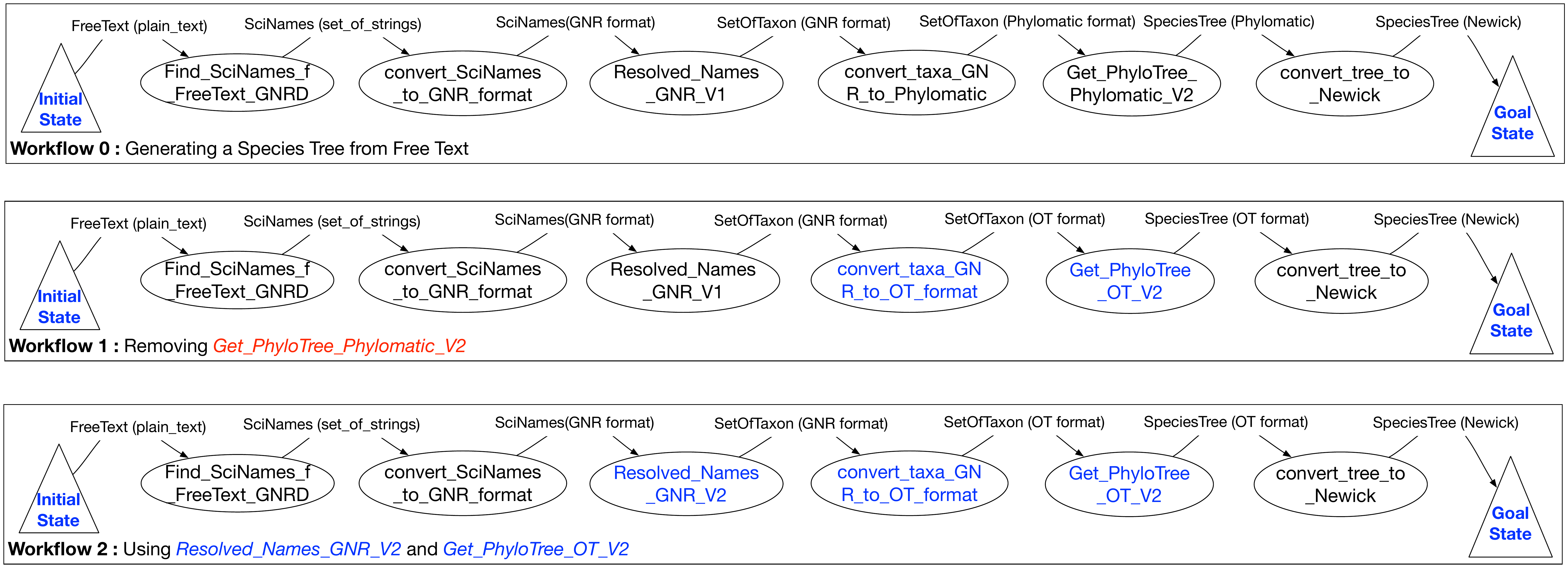}
	\caption{Generating a Species Tree from Free Text  \label{user_case_2_experiement}}
\end{figure}

The second use case is concerned with the generation of a species tree in  \texttt{\small newick} format from a text document  (\texttt{\small plain\_text} format). Figure \ref{user_case_2_experiement} (Workflow 0) displays the workflow with the highest QoS value. The first revision that an user asked is the system is to avoid using the \texttt{\small Get\_PhyloTree\_Phylomatic\_V2} service. The resulting most similar workflow is displayed in 
Figure \ref{user_case_2_experiement} (Workflow 1) in which the service is replaced by the service \texttt{\small Get\_PhyloTree\_OT\_V2}. However, for this service to be used, the service \texttt{\small convert\_taxa\_GNR\_to\_Phylomatic} must be replaced with \texttt{\small convert\_taxa\_GNR\_to\_OT\_format}. The second revision was to force the use of the service  \texttt{\small Resolved\_Names\_GNR\_V2}, the third workflow Figure \ref{user_case_2_experiement} (Workflow 2) satisfies this request. 

\subsubsection{Generating a Chronogram from a Web-site Content}

\begin{figure}[t]
	\centering	\includegraphics[width=\textwidth]{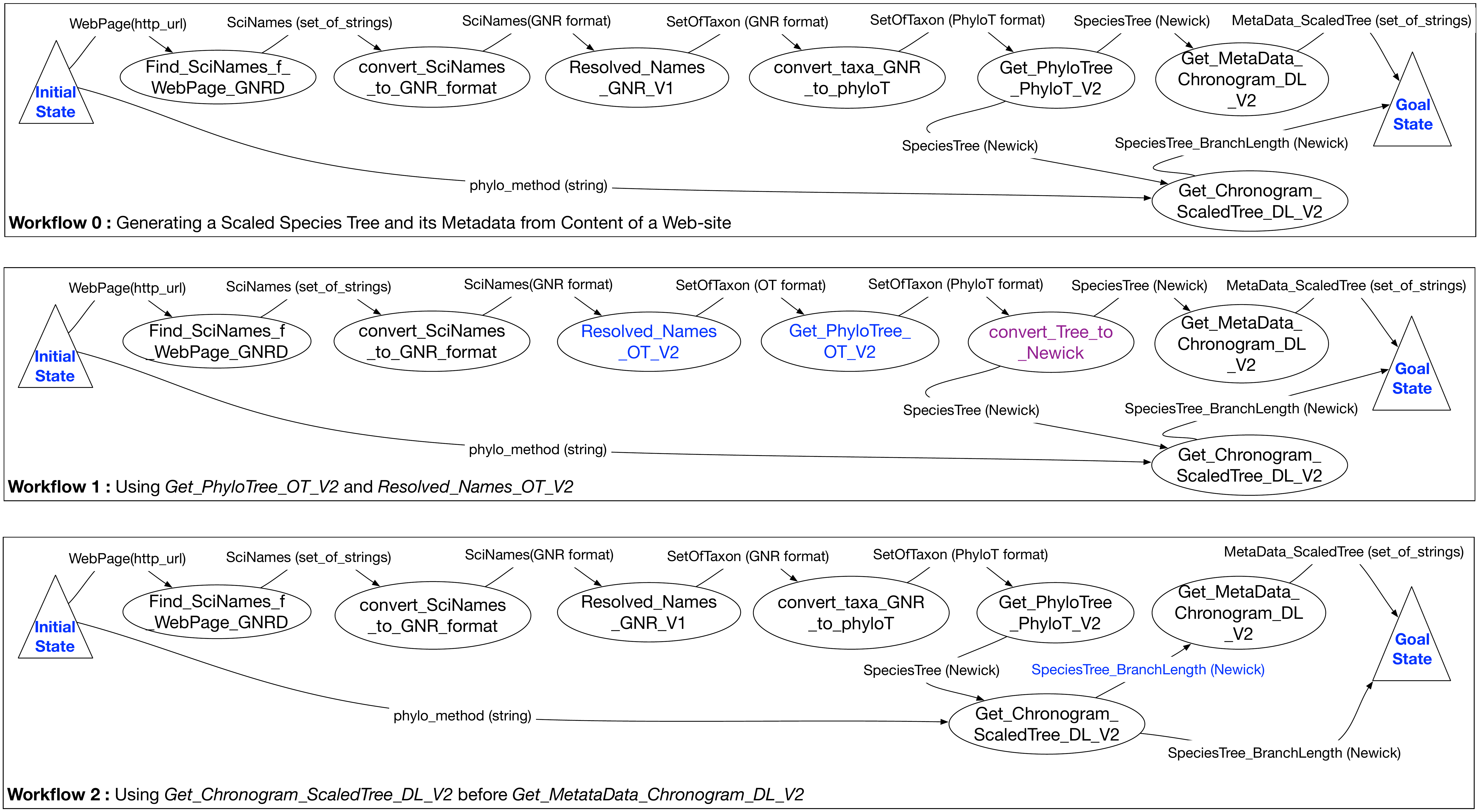}
	\caption{Generating a Scaled Species Tree and its Metadata from Content of a Web-site \label{user_case_3_experiement}}
\end{figure}

A \texttt{\small chronogram} is a scaled species tree with branch lengths. In this use case, 

\begin{list}{$\bullet$}{\itemsep=0pt \parsep=1pt \topsep=1pt} 

\item An user has an initial request to create a chronogram in the \texttt{\small newick} format and meta-data of this tree in the \texttt{\small set\_of\_strings} format from a web-site content, which is specified by a URL (\texttt{\small http\_url} format), and a name of scaling method (\texttt{\small string} format). The workflow  generated by the planning engine is shown in Figure \ref{user_case_3_experiement} (Workflow 0).
In this workflow, the output components \texttt{\small chronogram} and \texttt{\small meta-data} are produced by services \texttt{\small Get\_Chronogram\_ScaledTree\_DL\_V2} and \texttt{\small Get\_MetaData\_Chronogram\_DL\_V2} respectively. Both of them use \texttt{\small species\_tree} as an input and this resource is generated by service \texttt{\small Get\_PhyloTree\_PhyloT\_V2} in previous step. 

\item The user then requests that the services \texttt{\small Get\_PhyloTree\_OT\_V2} and \texttt{\small Resolved\_Names\_OT\_V2} have to be used.  
The resulting most similar workflow is presented in Figure \ref{user_case_3_experiement} (Workflow 1) in which services \texttt{\small Resolved\_Names\_GNR\_V1}, \texttt{\small convert\_taxa\_GNR\_to\_phyloT} and \texttt{\small Get\_PhyloTree\_PhyloT\_V2} are replaced by \texttt{\small Resolved\_Names\_OT\_V2}, \texttt{\small Get\_PhyloTree\_OT\_V2} and \texttt{\small convert\_Tree\_to\_Newick} respectively. 

\item Instead of the above request, the user requests that \texttt{\small Get\_MetaData\_Chronogram\_DL\_V2} is executed \textbf{after}   \texttt{\small Get\_Chronogram\_ScaledTree\_DL\_V2}. 
Figure \ref{user_case_3_experiement} (Workflow 2) shows the most similar workflow to Workflow 0 that satisfies this request. 
In this workflow,  \texttt{\small Get\_MetaData\_Chronogram\_DL\_V2} will use the output of 
\texttt{\small Get\_Chronogram\_ScaledTree\_DL\_V2} as its input instead of the output from  \texttt{\small Get\_PhyloTree\_PhyloT\_V2}.


\end{list}

\section{Conclusion and Future Works}

In this paper, we described two enhancements to the Phylotastic project 
that allow users to select their most preferred workflow and modify a
workflow and obtain the most similar workflow to the original one. 
In the process, we defined the notion of quality of service of a workflow 
and the notion of similarity between two workflows. We discuss their 
implementation and their use in enhancing the capabilities of the Phylotastic system. 
The proposed system is currently begin evaluated by biology researchers participating to the Phylotastic 
project. Our immediate future considerations are: {\em (i)} investigate the use of node QoS or similarity
in the ASP-planning engine to assist the computation of most preferred (or similar) workflow;
{\em (ii)} study other extensions of {\bf clingo} (e.g., clingcon) that allow a tighter integration of 
CSP and ASP in computing most preferred (or similar) workflows; 
{\em (iii)} evaluate the scalability and efficiency of the system when more web services are 
registered to Phylotastic.

\bibliographystyle{acmtrans}
\bibliography{./enrico,./bibfile,./bib2010,./tnguyen2018}

\begin{thebibliography}{}

\bibitem[\protect\citeauthoryear{{A. Stoltzfus et al.}}{{A. Stoltzfus et
  al.}}{2013}]{phylotastic1}
{\sc {A. Stoltzfus et al.}} 2013.
\newblock {Phylotastic! Making tree-of-life knowledge accessible, reusable and
  convenient}.
\newblock {\em BMC Bioinformatics\/}~{\em 14}.

\bibitem[\protect\citeauthoryear{Antunes, Bakhshandeh, Borbinha, Cardoso,
  Dadashnia, Francescomarino, Dragoni, Fettke, Gal, Ghidini, Hake, Khiat,
  Klinkm{\"{u}}ller, Kuss, Leopold, Loos, Meilicke, Niesen, Pesquita,
  P{\'{e}}us, Schoknecht, Sheetrit, Sonntag, Stuckenschmidt, Thaler, Weber, and
  Weidlich}{Antunes et~al\mbox{.}}{2015}]{AntunesBBCDFDFG15}
{\sc Antunes, G.}, {\sc Bakhshandeh, M.}, {\sc Borbinha, J.~L.}, {\sc Cardoso,
  J.}, {\sc Dadashnia, S.}, {\sc Francescomarino, C.~D.}, {\sc Dragoni, M.},
  {\sc Fettke, P.}, {\sc Gal, A.}, {\sc Ghidini, C.}, {\sc Hake, P.}, {\sc
  Khiat, A.}, {\sc Klinkm{\"{u}}ller, C.}, {\sc Kuss, E.}, {\sc Leopold, H.},
  {\sc Loos, P.}, {\sc Meilicke, C.}, {\sc Niesen, T.}, {\sc Pesquita, C.},
  {\sc P{\'{e}}us, T.}, {\sc Schoknecht, A.}, {\sc Sheetrit, E.}, {\sc Sonntag,
  A.}, {\sc Stuckenschmidt, H.}, {\sc Thaler, T.}, {\sc Weber, I.}, {\sc and}
  {\sc Weidlich, M.} 2015.
\newblock The process model matching contest 2015.
\newblock In {\em Enterprise Modelling and Information Systems Architectures,
  Proceedings of the 6th Int. Workshop on Enterprise Modelling and Information
  Systems Architectures, {EMISA} 2015, Innsbruck, Austria, September 3-4,
  2015.} 127--155.

\bibitem[\protect\citeauthoryear{Becker and Laue}{Becker and
  Laue}{2012}]{BeckerL12}
{\sc Becker, M.} {\sc and} {\sc Laue, R.} 2012.
\newblock A comparative survey of business process similarity measures.
\newblock {\em Computers in Industry\/}~{\em 63,\/}~2, 148--167.

\bibitem[\protect\citeauthoryear{Bininda-Emonds, Cardillo, Jones, MacPhee,
  Beck, Grenyer, Price, Vos, Gittleman, and Purvis}{Bininda-Emonds
  et~al\mbox{.}}{2007}]{ph4}
{\sc Bininda-Emonds, O.}, {\sc Cardillo, M.}, {\sc Jones, K.}, {\sc MacPhee,
  R.}, {\sc Beck, R.}, {\sc Grenyer, R.}, {\sc Price, S.}, {\sc Vos, R.}, {\sc
  Gittleman, J.}, {\sc and} {\sc Purvis, A.} 2007.
\newblock {The delayed rise of present-day mammals}.
\newblock {\em Nature\/}~{\em 446,\/}~7135.

\bibitem[\protect\citeauthoryear{Carman, Serafini, and Traverso}{Carman
  et~al\mbox{.}}{2004}]{compose4}
{\sc Carman, M.}, {\sc Serafini, L.}, {\sc and} {\sc Traverso, P.} 2004.
\newblock {Web Service Composition as Planning}.
\newblock In {\em Proceedings of ICAPS 2003 Workshop on Planning for Web
  Services}.

\bibitem[\protect\citeauthoryear{Cracraft, Donoghue, Dragoo, Hillis, and
  Yates}{Cracraft et~al\mbox{.}}{2002}]{ph2}
{\sc Cracraft, J.}, {\sc Donoghue, M.}, {\sc Dragoo, J.}, {\sc Hillis, D.},
  {\sc and} {\sc Yates, T.} 2002.
\newblock {Assembling the tree of life: harnessing life's history to benefit
  science and society}.
\newblock Tech. Rep. \url{http://ucjeps.berkeley.edu/tol.pdf.}, U.C. Berkeley.

\bibitem[\protect\citeauthoryear{Jetz, Thomas, Joy, Hartmann, and Mooers}{Jetz
  et~al\mbox{.}}{2012}]{birds}
{\sc Jetz, W.}, {\sc Thomas, G.}, {\sc Joy, J.}, {\sc Hartmann, K.}, {\sc and}
  {\sc Mooers, A.} 2012.
\newblock The global diversity of birds in space and time.
\newblock {\em Nature\/}~{\em 491,\/}~7424.

\bibitem[\protect\citeauthoryear{Kaminski, Schaub, and Wanko}{Kaminski
  et~al\mbox{.}}{2017}]{KaminskiSW17}
{\sc Kaminski, R.}, {\sc Schaub, T.}, {\sc and} {\sc Wanko, P.} 2017.
\newblock A tutorial on hybrid answer set solving with clingo.
\newblock In {\em Reasoning Web. Semantic Interoperability on the Web - 13th
  International Summer School 2017, London, UK, July 7-11, 2017, Tutorial
  Lectures}. 167--203.

\bibitem[\protect\citeauthoryear{Lifschitz}{Lifschitz}{2002}]{Lifschitz02}
{\sc Lifschitz, V.} 2002.
\newblock {Answer set programming and plan generation}.
\newblock {\em Artificial Intelligence\/}~{\em 138,\/}~1--2, 39--54.

\bibitem[\protect\citeauthoryear{McIlraith, Son, and Zeng}{McIlraith
  et~al\mbox{.}}{2001}]{mcisonzen01}
{\sc McIlraith, S.}, {\sc Son, T.}, {\sc and} {\sc Zeng, H.} 2001.
\newblock Semantic {W}eb services.
\newblock {\em IEEE Intelligent Systems (Special Issue on the Semantic
  {W}eb)\/}~{\em 16,\/}~2 (March/April), 46--53.

\bibitem[\protect\citeauthoryear{Mora, Tittensor, Adl, Simpson, and Worm}{Mora
  et~al\mbox{.}}{2011}]{ph1}
{\sc Mora, C.}, {\sc Tittensor, D.}, {\sc Adl, S.}, {\sc Simpson, A.}, {\sc
  and} {\sc Worm, B.} 2011.
\newblock {How many species are there on Earth and in the ocean?}
\newblock {\em {PLoS biology}\/}~{\em 9,\/}~8.

\bibitem[\protect\citeauthoryear{Nguyen, Son, and Pontelli}{Nguyen
  et~al\mbox{.}}{2018}]{NguyenSP18}
{\sc Nguyen, T.~H.}, {\sc Son, T.~C.}, {\sc and} {\sc Pontelli, E.} 2018.
\newblock Automatic web services composition for phylotastic.
\newblock In {\em Practical Aspects of Declarative Languages - 20th
  International Symposium}. 186--202.

\bibitem[\protect\citeauthoryear{Peer}{Peer}{2005}]{compose7}
{\sc Peer, J.} 2005.
\newblock {Web Service Composition as AI Planning - a Survey}.
\newblock Tech. rep., University of St. Gallen.

\bibitem[\protect\citeauthoryear{Penny}{Penny}{2004}]{felse}
{\sc Penny, D.} 2004.
\newblock Inferring phylogenies.—joseph felsenstein. 2003. sinauer
  associates, sunderland, massachusetts.
\newblock {\em Systematic Biology\/}~{\em 53,\/}~4, 669--670.

\bibitem[\protect\citeauthoryear{Prosdocimi, Chisham, Thompson, Pontelli, and
  Stoltzfus}{Prosdocimi et~al\mbox{.}}{2009}]{cdao1}
{\sc Prosdocimi, F.}, {\sc Chisham, B.}, {\sc Thompson, J.}, {\sc Pontelli,
  E.}, {\sc and} {\sc Stoltzfus, A.} 2009.
\newblock Initial implementation of a comparative data analysis ontology.
\newblock {\em Evolutionary Bioinformatics\/}~{\em 5}, 47--66.

\bibitem[\protect\citeauthoryear{Rajendran and Balasubramanie}{Rajendran and
  Balasubramanie}{2009}]{RajendranB09}
{\sc Rajendran, T.} {\sc and} {\sc Balasubramanie, D.~P.} 2009.
\newblock Analysis on the study of qos-aware web services discovery.
\newblock {\em Journal of Computing\/}, 119--130.

\bibitem[\protect\citeauthoryear{Smith, Beaulieu, Stamatakis, and
  Donoghue}{Smith et~al\mbox{.}}{2011}]{ph5}
{\sc Smith, S.}, {\sc Beaulieu, J.}, {\sc Stamatakis, A.}, {\sc and} {\sc
  Donoghue, M.} 2011.
\newblock {Understanding angiosperm diversification using small and large
  phylogenetic trees}.
\newblock {\em American Journal of Botology\/}~{\em 98,\/}~3, 404--414.

\bibitem[\protect\citeauthoryear{Stoltzfus, O'Meara, Whitacre, Mounce,
  Gillespie, Kumar, Rosauer, and Vos}{Stoltzfus et~al\mbox{.}}{2012}]{ph3}
{\sc Stoltzfus, A.}, {\sc O'Meara, B.}, {\sc Whitacre, J.}, {\sc Mounce, R.},
  {\sc Gillespie, E.}, {\sc Kumar, S.}, {\sc Rosauer, D.}, {\sc and} {\sc Vos,
  R.} 2012.
\newblock {Sharing and Re-use of Phylogenetic Trees (and associated data) to
  Facilitate Synthesis}.
\newblock {\em BMC Research Notes\/}~{\em 5}, 574.

\bibitem[\protect\citeauthoryear{Zhang and Shasha}{Zhang and
  Shasha}{1989}]{ZhangS89}
{\sc Zhang, K.} {\sc and} {\sc Shasha, D.~E.} 1989.
\newblock Simple fast algorithms for the editing distance between trees and
  related problems.
\newblock {\em {SIAM} J. Comput.\/}~{\em 18,\/}~6, 1245--1262.

\end{thebibliography}
\label{lastpage}
\end{document}